\documentclass[conference]{IEEEtran}
\IEEEoverridecommandlockouts
\usepackage{amsmath,amssymb,amsfonts}
\usepackage{algorithmic}
\usepackage{graphicx}
\usepackage{textcomp}
\usepackage{xcolor}
\usepackage{hyperref}
\usepackage{multirow}

\usepackage[backend=biber,style=ieee]{biblatex}
\def\BibTeX{{\rm B\kern-.05em{\sc i\kern-.025em b}\kern-.08em
    T\kern-.1667em\lower.7ex\hbox{E}\kern-.125emX}}

\begin{document}

\title{\MakeUppercase{
Inner Loop Inference for Pretrained Transformers:
Unlocking Latent Capabilities Without Training
}}

\author{\IEEEauthorblockN{Jonathan Lys, Vincent Gripon,\\Bastien Pasdeloup, Axel Marmoret}
\\
\IEEEauthorblockA{IMT Atlantique, Lab-STICC,\\ UMR CNRS 6285, Brest, France \\
\href{mailto:name.surname@imt-atlantique.fr}{\texttt{name.surname@imt-atlantique.fr}}}
\and
\IEEEauthorblockN{Lukas Mauch, Fabien Cardinaux,\\Ghouthi Boukli Hacene}
\\
\IEEEauthorblockA{Sony Europe Ltd. \\ 
Stuttgart Technology Center, EUREC, Germany
\\
\href{mailto:Name.Surname@sony.com}{\texttt{Name.Surname@sony.com}}}
}

\maketitle

\begin{abstract}
Deep Learning architectures, and in particular Transformers, are conventionally viewed as a composition of layers. These layers are actually often obtained as the sum of two contributions: a residual path that copies the input and the output of a Transformer block.
As a consequence, the inner representations (i.e. the input of these blocks) can be interpreted as iterative refinement of a propagated latent representation. Under this lens, many works suggest that the inner space is shared across layers, meaning that tokens can be decoded at early stages. Mechanistic interpretability even goes further by conjecturing that some layers act as refinement layers. Following this path, we propose \emph{inference-time inner looping}, which prolongs refinement in pretrained off-the-shelf language models by repeatedly re-applying a selected block range.
Across multiple benchmarks, inner looping yields modest but consistent accuracy improvements. Analyses of the resulting latent trajectories suggest more stable state evolution and continued semantic refinement. Overall, our results suggest that additional refinement can be obtained through simple test-time looping, extending computation in frozen pretrained models.

\end{abstract}


\section{Introduction}

Modern language models are built on Transformer architectures~\cite{vaswani_attention_2017} that process complex context to generate text. Central to this design are residual connections, which propagate hidden states throughout the network's entire depth. Some authors suggest that these connections allow each layer to incrementally refine the representation of the predicted token~\cite{lad2024the}. From this perspective, next-token prediction can be viewed as a sequence of iterative context-conditioned state updates over a common state vector.

This refinement view aligns with mechanistic interpretability~\cite{saphra2024mechanistic}, where understanding model behavior relies on decomposing the incremental contributions of individual layers. Specifically, logit-lens readouts show that intermediate residual states already decode into meaningful “proto-predictions” that are typically sharpened rather than replaced later in the network~\cite{nostalgebraist2020logitlens,belrose2023eliciting}. Robustness studies further suggest that many layers can be removed or permuted with only limited accuracy degradation, pointing to redundancy in evidence accumulation and arguing against a brittle stage-by-stage pipeline~\cite{lad2024the}. At the circuit level, induction heads provide a concrete example of depth-wise state evolution: early layers write retrieval cues that later layers resolve, effectively unrolling a retrieval-and-copy computation over depth~\cite{olsson2022context}. 
We summarize these observations as a \emph{logits refinement} view of depth: depth primarily governs how many refinement steps are applied to a shared latent state. 

If depth acts as a number of refinement updates applied to a state transmitted across layers, a natural question follows: \emph{is it possible to add refinement steps at inference time, improving the model performance, without changing model parameters?} 

We investigate this via \emph{inference-time inner looping}: we repeatedly re-apply a chosen range of layers to the same hidden state, using a lightweight interpolation mechanism to stabilize the resulting trajectory. This enables the accumulation of additional refinement steps while keeping the latent state close to the model’s standard forward pass.

This investigation parallels recent architectural approaches that explicitly reuse computation in latent space, often referred to as latent reasoning. For example, CoCoNut~\cite{hao2025training} replaces explicit chain-of-thought traces with continuous latent tokens that are re-injected into the context, treating the hidden state as a recurrent workspace during training. Huginn~\cite{geiping2025scaling} similarly demonstrates that \emph{middle looping}, the deliberate re-utilization of a subset of central layers, can approximate increased depth and improve reasoning capability without increasing parameter count. Related work further suggests that repeatedly reusing layer-level circuits induces a bias toward iterative computation and reasoning~\cite{saunshi2025reasoning}. In contrast, our focus is not on training new architectures or scaling strategies, but on using looping as a \emph{test-time probe} of refinement dynamics in \emph{frozen} pretrained models, uncovering an intrinsic hidden potential.

We evaluate inner looping across multiple benchmarks and analyze the resulting latent trajectories. We find that adding regularized refinement steps yields modest but consistent accuracy improvements, and that the induced representation trajectories are indicative of continued semantic disambiguation and occasional self-correction. Those results are consistent with the logits refinement hypothesis.

\noindent\textbf{Contributions.} This paper makes the following contributions:
\begin{itemize}
    \item We propose a simple framework for \emph{inference-time inner looping} in off-the-shelf language models, extending computation by re-applying a selected block range.
    \item We introduce \emph{representation interpolation} as a lightweight regularization mechanism that keeps looped states close to their non-looped counterparts to reduce distribution shift.
    \item We provide empirical evidence across multiple benchmarks that inner looping yields modest but consistent accuracy gains in frozen pretrained models.
\end{itemize}

\section{Methodology}

\subsection{Naive Looping}

We first establish a formal framework for looping. We define standard Transformer notations and introduce our modification: a mechanism to decouple layer execution from network depth. A Transformer consists of an embedding layer $E$ that maps a sequence of token indices to initial hidden states:
\begin{equation}
h_0 = \big( E[x_1], \dots, E[x_T]\big) \;, 
\quad E \in \mathbb{R}^{|V|\times d} \;.
\end{equation}

A fixed set of Transformer blocks is then applied:
\begin{equation}
h_\ell = \mathcal{B}_\ell(h_{\ell-1}) \;, \quad \ell = 1,\dots,L \;,
\end{equation}
where each block $\mathcal{B}_\ell$ consists of multi-head self-attention (MSA) and a feed-forward network (FFN), combined with residual connections and layer normalization (LN):
\begin{equation}
z = h_{\ell-1} + \mathrm{MSA}_\ell\!\big(\mathrm{LN}(h_{\ell-1})\big) \;,
\end{equation}
\begin{equation}
h_\ell = z + \mathrm{FFN}_\ell\!\big(\mathrm{LN}(z)\big) \;.
\end{equation}

In the case of Gemma, additional post-normalization layers are applied. At the end of the network, a final normalization layer and an unembedding matrix map the hidden states to token probabilities.

Inner looping decouples the hidden-state index from the block index, allowing selected blocks to be applied multiple times. We introduce a step index $k \in \{0,\dots,K-1\}$ for hidden states, with $K > L$, and a block mapping $\pi : \{0,\dots,K-1\} \to \{0,\dots,L-1\}$ specifying which block is applied at each step.

We adopt the \emph{middle-looping} scheme of~\cite{saunshi2025reasoning, geiping2025scaling}, defined by three parameters:

\begin{itemize}
    \item $s$: index of the first block in the loop,
    \item $e$: index of the block following the loop,
    \item $R$: number of repetitions.
\end{itemize}

Blocks before $s$ are applied once, blocks in $\{s, s+1, \dots, e-1\}$ are repeated $R$ times, and blocks after $e$ are applied once. Let $N = e-s$ denote the loop length and $K = L + (R-1)N$ the total number of block applications. The corresponding mapping $\pi:\{1,\dots,K\}\to\{1,\dots,L\}$ is:
\begin{equation}
\pi(k) = 
\begin{cases}
k & k < s \\
s + (k-s)\bmod N & s \leq k < s + RN \\
k - (R-1)N & k \geq s + RN
\end{cases} \;.
\end{equation}

However, simply repeating blocks alters both the effective depth and the scale of the latent state. We hypothesize that this naive application induces a distribution shift, pushing activations away from the manifold encountered during standard inference. In Section~\ref{subsec:naive}, we empirically test this hypothesis and show that uncontrolled looping indeed leads to systematic degradation consistent with such a shift.

\subsection{Regularized Looping}
\label{sec:reg_looping}

To address this potential mismatch, we introduce \emph{regularized looping}. This approach explicitly interpolates the looped hidden states with a baseline (non-looped) reference, ensuring that the intermediate representations remain grounded within the model's valid activation distribution.

To simplify notation, we describe the update at the loop boundary and omit depth indices in this subsection. Let $h^{(0)}$ denote the baseline hidden state at the loop boundary obtained from a standard forward pass. Let $F$ denote one loop application (\emph{e.g.}, applying the middle block range once). At loop step $t\ge 1$, we first compute an unregularized loop output:
\begin{equation}
h^{(t)} = F\!\left(\hat h^{(t-1)}\right) \;,
\end{equation}
and then form a regularized state $\hat h^{(t)}$ from a cache of past loop states $\{h^{(i)}\}_{i=0}^{t}$:
\begin{equation}
\hat h^{(t)} = \sum_{i=0}^{t} \alpha_i\, h^{(i)} \;,
\text{ subject to } \sum_{i=0}^{t}\alpha_i = 1 \;.
\end{equation}
The interpolation methods differ in their weighting coefficients $\alpha_i$. Each strategy encodes a different prior over the cached states.

\paragraph{Naive looping (special case)}
Naive looping omits the caching mechanism, \emph{i.e.}, $\alpha_t=1$ and $\alpha_{i<t}=0$. In this case, $\hat h^{(t)} = h^{(t)}$ and the recursion reduces to:
\begin{equation}
h^{(t)} = F\!\left(h^{(t-1)}\right) \;, \qquad t=1,\dots,R \;.
\end{equation}

\paragraph{Uniform}
Average all cached states:
\begin{equation}
\hat h^{(t)} = \frac{1}{t+1}\sum_{i=0}^{t} h^{(i)} \;.
\end{equation}

\paragraph{Moving average}
Interpolate the current loop output with the baseline value:
\begin{equation}
\hat h^{(t)} = \eta \cdot h^{(0)} + (1-\eta)\cdot h^{(t)} \;.
\end{equation}

\paragraph{Auto-alignment}
We compute a score vector by aligning each cached state with the baseline state, and apply a softmax over these scores:
\begin{equation}
s^{(t)} = \Big[\langle h^{(0)}, h^{(0)}\rangle,\dots,\langle h^{(0)}, h^{(t)}\rangle\Big]\in\mathbb{R}^{t+1} \;,
\end{equation}
\begin{equation}
\alpha^{(t)} = \mathrm{softmax}\!\left(s^{(t)}\right)\in\mathbb{R}^{t+1} \;,
\quad
\hat h^{(t)} = \sum_{i=0}^{t} \alpha^{(t)}_i\, h^{(i)} \;.
\end{equation}

\section{Experimental setup}

\subsection{Model}

In our experiments, we evaluate inner looping on two pretrained model families: Gemma~2 (2B and 9B)~\cite{team2024gemma} and Llama~3–8B~\cite{grattafiori2024llama}. These models primarily differ in their normalization schemes, enabling us to examine how architectural design influences the stability of iterative inference.

Although the looping algorithm itself is generic, its empirical behavior could strongly depend on the underlying Transformer architecture. When inference is treated as an iterative dynamical process, seemingly minor design choices, such as normalization placement, can substantially affect stability under repeated residual updates.
The original Transformer~\cite{vaswani_attention_2017} used post-layer normalization, a design also adopted in iterative architectures such as the Universal Transformer~\cite{dehghani2018universal}. More recent hybrid schemes (\emph{e.g.}, “sandwich” normalization~\cite{geiping2025scaling}) further target stability under repeated computation. In contrast, most modern LLMs rely on pre-layer normalization, which improves optimization during training but is known to be less stable when blocks are repeatedly applied, particularly in implicit regimes~\cite{bai2021stabilizing}.

Gemma~2 integrates both pre- and post-normalization within each block, offering partial structural support for iterative refinement. Llama~3–8B, by comparison, follows a standard pre-norm design. This contrast allows us to assess whether inference-time regularization can mitigate the instability typically associated with pre-norm architectures, and more generally, whether inner looping extends beyond models that implicitly favor iterative computation.

\subsection{Datasets}

We provide an overview of the evaluation datasets used to assess the performance of our method. Most benchmarks are multiple-choice, with a single correct answer. For each trial, the model computes the cumulative probability of every option, the prediction is counted as correct if the true answer achieves the highest length-normalized likelihood.

Some benchmarks use targeted examples, or \emph{shots}, to provide context for the questions, with the number of shots varying from 0 to 25. Collectively, these datasets evaluate both the models’ reasoning capabilities and their internal knowledge. We provide a short description of each dataset:

\begin{itemize}
\item \textbf{WinoGrande}~\cite{winogrande2021}: A dataset designed to test commonsense reasoning via pronoun resolution. Each sample consists of a sentence with an ambiguous pronoun, requiring models to resolve the correct referent based on context. 

\item \textbf{ARC (AI2 Reasoning Challenge) -- Easy and Challenge}~\cite{arc2018}: The ARC dataset presents grade-school science questions with multiple-choice answers, separated into “Easy” and “Challenge” subsets.

\item \textbf{GSM8K}~\cite{cobbe2021gsm8k}: A generative benchmark for grade-school math word problems. Each problem requires multi-step quantitative reasoning and arithmetic.

\item \textbf{HellaSwag}~\cite{zellers2019hellaswag}: A commonsense inference benchmark consisting of multiple-choice questions. Each instance is a context followed by four possible sentence continuations, with the task of selecting the most plausible ending.

\item \textbf{MMLU (Massive Multitask Language Understanding)}~\cite{hendrycks2021measuring}: A comprehensive benchmark covering 57 tasks across subjects such as mathematics, history, science, law, medicine, and more.

\end{itemize}

To ensure reproducibility, we rely on the lighteval library~\cite{lighteval}, which is built upon the HELM framework~\cite{liang2023holistic}. Code to reproduce experiments can be found on GitHub\footnote{\href{https://github.com/jonathanlys01/looped-transformer}{github.com/jonathanlys01/looped-transformer}}.

\section{Experimental Results}
\label{sec:results}

\begin{figure}[ht!]
    \centering
    \includegraphics[width=\linewidth]{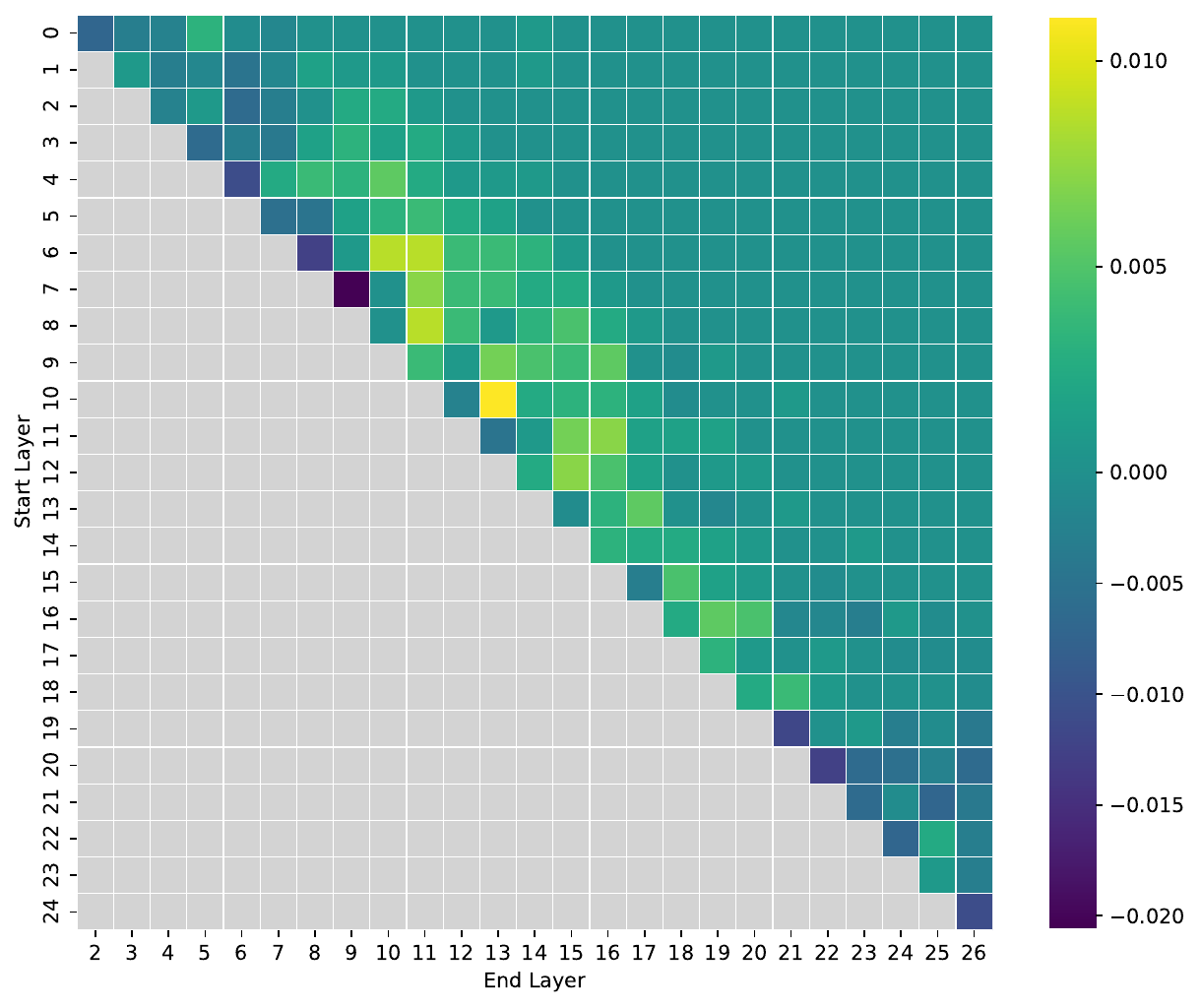}
    \caption{Heatmap of the accuracy difference relative to the baseline ($0.6875$) across start–end layer configurations under uniform interpolation.}
    \label{fig:uniform}
\end{figure}

\begin{table*}[ht!]
    \caption{Accuracy of the three regularized looping methods and the noise ablation across all benchmarks. \\
    Best results are shown in \textbf{bold}. \underline{Underlined} values match or exceed the corresponding baseline.}

    \renewcommand{\arraystretch}{1.9}
    \centering
    \begin{tabular}{|c|c|c|c|c|c|c|c|}
    \hline
        \textbf{Benchmark (shots)}  &\textbf{Model}&\textbf{Size}& \textbf{Baseline} & \textbf{Uniform} & \textbf{Moving Average} & \textbf{Auto-Align} & \textbf{Noise} \\ \hline
        \multirow{3}{*}{\textbf{WinoGrande (5)}}
        &\multirow{2}{*}{Gemma}&2B& 68.75 ± 1.30&\underline{\textbf{69.53 ± 1.29}}& \underline{69.06 ± 1.30}& \underline{68.98 ± 1.30}& 68.51 ± 1.31\\\cline{3-8}
        & &9B& 77.35 ± 1.18& \underline{\textbf{77.43 ± 1.17}}& \underline{77.35 ± 1.18}& \underline{77.35 ± 1.18}& \underline{77.35 ± 1.18}\\\cline{2-8}
        & Llama &8B& 76.16 ± 1.20& \underline{\textbf{76.64 ± 1.19}}& \underline{76.40 ± 1.19}& 76.09 ± 1.20& 75.93 ± 1.20\\ \hline
        \multirow{3}{*}{\textbf{Arc E (0)}}
        &\multirow{2}{*}{Gemma}&2B& \textbf{80.30 ± 0.82}& 80.01 ± 0.82& 80.18 ± 0.82& 80.22 ± 0.82& 79.84 ± 0.82\\ \cline{3-8}
        & &9B& 87.84 ± 0.67& \underline{87.84 ± 0.67}& \underline{\textbf{87.88 ± 0.67}}& \underline{87.84 ± 0.67}& \underline{87.84 ± 0.67}\\\cline{2-8}
        & Llama  &8B& \textbf{77.69 ± 0.85} & 77.31 ± 0.86& 77.40 ± 0.86& 77.61 ± 0.86& 77.57 ± 0.86\\ \hline
        \multirow{3}{*}{\textbf{Arc C (25)}}
        &\multirow{2}{*}{Gemma}&2B& 52.82 ± 1.46& \underline{\textbf{53.41 ± 1.46}}& \underline{52.90 ± 1.46}& \underline{52.99 ± 1.46}& 52.65 ± 1.46\\ \cline{3-8}
        & &9B& 67.75 ± 1.37& \underline{\textbf{68.09 ± 1.36}}& \underline{\textbf{68.09 ± 1.36}}& 67.58 ± 1.37&67.49 ± 1.37\\\cline{2-8}
        & Llama  &8B& 59.73 ± 1.43& 59.64 ± 1.43& \underline{59.73 ± 1.43}& \underline{\textbf{60.07 ± 1.43}}& 59.64 ± 1.43\\ \hline
        \multirow{2}{*}{\textbf{GSM8K (5)}}
        &\multirow{2}{*}{Gemma}&2B& 25.02 ± 1.19& \underline{26.00 ± 1.21}& \underline{\textbf{26.38 ± 1.21}}& \underline{26.08 ± 1.21}& 24.79 ± 1.19\\ \cline{3-8}
        & &9B& 68.46 ± 1.28& \underline{\textbf{68.52 ± 1.27}}& 68.39 ± 1.28& 68.31 ± 1.28& 68.33 ± 1.28\\\hline
        \multirow{3}{*}{\textbf{HellaSwag (10)}}
        &\multirow{2}{*}{Gemma}&2B& 74.51 ± 0.43& \underline{\textbf{74.70 ± 0.43}}& \underline{74.64 ± 0.43}& \underline{74.60 ± 0.43}& 74.50 ± 0.43\\ \cline{3-8}
        & &9B& 82.41 ± 0.38& \underline{\textbf{82.53 ± 0.38}}& \underline{82.42 ± 0.38}& \underline{82.42 ± 0.38}& \underline{82.42 ± 0.38}\\\cline{2-8}
        & Llama  &8B& 82.14 ± 0.38& \underline{82.30 ± 0.38}& \underline{\textbf{82.31 ± 0.38}}& \underline{\textbf{82.31 ± 0.38}}& \underline{82.14 ± 0.38}\\ \hline
        \multirow{3}{*}{\textbf{MMLU (5)}}
        &\multirow{2}{*}{Gemma}&2B& 53.93 ± 3.57& \underline{\textbf{54.49 ± 3.58}}& \underline{54.35 ± 3.57}& \underline{54.13 ± 3.57}& \underline{53.97 ± 3.58}\\\cline{3-8}
        & &9B& 72.17 ± 3.11& \underline{\textbf{72.42 ± 3.10}}& \underline{72.23 ± 3.11}& \underline{72.18 ± 3.11}& \underline{72.20 ± 3.11}\\\cline{2-8}
        & Llama  &8B& \textbf{34.50 ± 3.47}& 33.95 ± 3.45& 33.99 ± 3.45& 34.06 ± 3.45& 33.74 ± 3.45\\ \hline
    \end{tabular}
    \label{tab:main}
\end{table*}

We explicitly validate the inner looping framework in three stages. First, we confirm the instability of naive looping to justify our regularization approach. We then identify optimal layer configurations and evaluate the method across reasoning benchmarks and analyze the resulting latent trajectories.

\subsection{Assessing the naive looping method}\label{subsec:naive}

We apply the naive looping strategy to the Gemma2-2B model on the WinoGrande benchmark, using a loop count of $3$ and sweeping over all start-end layer pairs. All looped configurations perform worse than the baseline accuracy, with some settings collapsing to near-chance performance. The full sweep results, including the corresponding heatmap over start–end configurations, are provided in the accompanying GitHub repository. We interpret this systematic degradation as evidence of a distribution shift: looping exposes layers to activation statistics that differ from those encountered during standard forward inference, for which the model was optimized. This mismatch appears sufficient to destabilize prediction quality. These observations motivate the regularization strategies introduced in Section~\ref{sec:reg_looping}.

\subsection{Layer Selection (ablation)}

We next rely on the uniform regularization strategy to select the loop start–end layers. For each model, we sweep all candidate layer intervals on WinoGrande. In the case of Gemma~2-2B, the best configuration corresponds to looping layers 10–13 (see Figure~\ref{fig:uniform}).
For other models, the selected intervals consistently fall within comparable relative depths, roughly 40–60\% of the network, matching the stage-wise inference dynamics reported in~\cite{lad2024the}. For subsequent benchmarks, we keep these model-specific loop regions fixed instead of re-optimizing them per task.

\subsection{Main Results}

In Table~\ref{tab:main}, we evaluate the regularized looping variants across multiple benchmarks, using for each model the loop configuration selected on WinoGrande. We report in columns "Uniform", "Moving Average" and "Auto-Align" all strategies introduced in Section~\ref{sec:reg_looping}.
Additionally, as a control, we replace the loop-induced signal with matched-magnitude random noise. This method is reported as "Noise".

For both Gemma models, inner looping improves performance on most reasoning benchmarks. The only exception is Gemma-2B on ARC Easy, evaluated in a zero-shot setting. Among the proposed regularizers, Uniform is the most stable and generally provides the largest gains. Taken together, these results suggest that Gemma models can benefit from additional, regularized refinement steps applied purely at inference time.

In contrast, Llama-3-8B shows a more variable response across tasks, indicating greater sensitivity to repeated iteration in this pre-norm architecture.

Finally, the noise ablation consistently underperforms the structured looping variants and frequently drops below the baseline. This supports the view that the observed improvements stem from coherent refinement dynamics rather than from perturbations of similar magnitude alone.

\subsection{Latent Trajectories Visualization}
\label{subsec:traj}

\begin{figure}[ht]
    \centering
    \includegraphics[width = .75\linewidth]{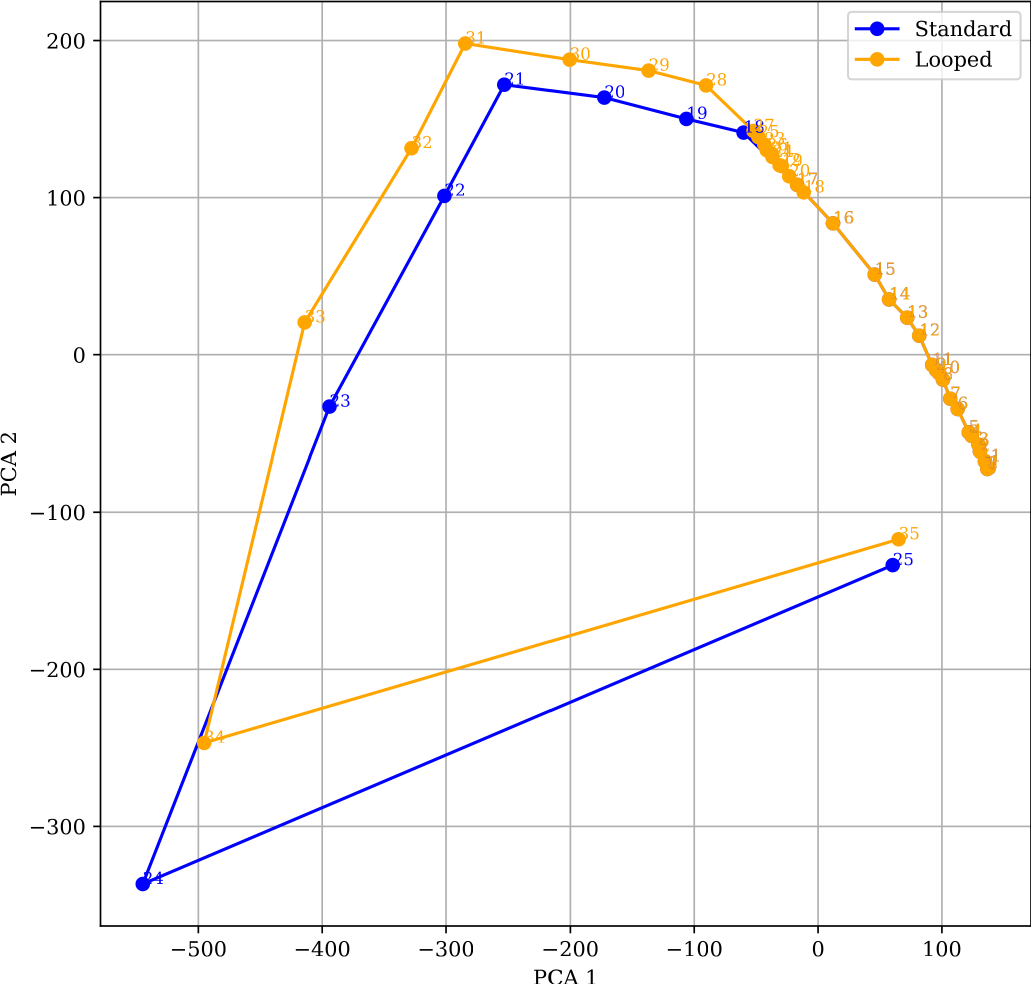}
    \caption{PCA projection of latent activation trajectories across model depth.}
    \label{fig:trajectories}
\end{figure}

Figure~\ref{fig:trajectories} visualizes hidden-state trajectories for the same input with and without looping, using a 2D PCA projection of the latent states for a randomly selected sample. During the looped segment, we observe a small deviation from the non-looped trajectory, indicating that additional iterations shift the representation away from its baseline path. This deviation then propagates through subsequent layers, yielding a slightly different output representation.

Notably, the overall trajectory structure remains similar between the two runs, suggesting that looping typically induces a small perturbation rather than a qualitatively different computation. A plausible interpretation is that such small shifts can be sufficient to change the final outcome by altering relative logit margins, which is consistent with the modest but consistent accuracy gains observed in Table~\ref{tab:main}.

\section{Conclusion}

In this work, we investigated whether the refinement dynamics suggested by mechanistic analyses of Transformers can be extended at inference time in \emph{frozen} pretrained models. We showed that naive inner looping is unstable, but that simple regularizers enable additional refinement steps without training. Across multiple benchmarks, this yields modest but consistent accuracy gains for Gemma models, while Llama~3--8B responds less reliably, due to architectural differences. Latent-trajectory visualizations are consistent with small, structured representation shifts that propagate to the output and occasionally align with self-correction. Overall, inference-time inner looping offers a lightweight way to expose depth-limited latent computation in off-the-shelf Transformers, further confirming the logits refinement hypothesis.

\section{Acknowledgements}

This research has been funded, in part, by the French National Research Agency under project ANR-24-CE23-7365. With a view to its publication in open access, the author has applied for an open access CC-BY licence for any manuscript accepted for publication resulting from this submission. This work was granted access to the HPC resources of IDRIS under the allocation 2024-AD011015938 made by GENCI.

\printbibliography

\end{document}